# Where you live matters: a spatial analysis of COVID-19 mortality

Dr Behzad Javaheri

Department of Computer Science, City University of London, London, UK. behzad.javaheri@city.ac.uk

**Abstract**— The COVID-19 pandemic has caused ~ 2 million fatalities. Significant progress has been made in advancing our understanding of the disease process, one of the unanswered questions, however, is the anomaly in the case/mortality ratio with Mexico as a clear example. Herein, this anomaly is explored by spatial analysis and whether mortality varies locally according to local factors. To address this, hexagonal cartogram maps (hexbin) used to spatially map COVID-19 mortality and visualise association with patient-level data on demographics and pre-existing health conditions. This was further interrogated at local Mexico City level by choropleth mapping. Our data show that the use of hexagonal cartograms is a better approach for spatial mapping of COVID-19 data in Mexico as it addresses bias in area size and population. We report sex/age-related spatial relationship with mortality amongst the Mexican states and a trend between health conditions and mortality at the state level. Within Mexico City, there is a clear south, north divide with higher mortality in the northern municipalities. Deceased patients in these northern municipalities have the highest pre-existing health conditions. Taken together, this study provides an improved presentation of COVID-19 mapping in Mexico and demonstrates spatial divergence of the mortality in Mexico.

———————————— ◆ ————————————

## 1 Problem Statement

It is more than 1 year since a new coronavirus strain, SARS-CoV-2, the underlying cause for the current pandemic has been identified [1]. Despite the global effort, COVID-19 has spread to 201 countries leading to ~90 million cases with ~2 million fatalities [2].

To better understand the disease aiming to suppress mortality, various aspects of the pandemic has been the subject of numerous studies covering disciplines including social sciences, medicine, mathematics and artificial intelligence.

A significant advance in our understanding of COVID-19 pandemic has been made, however, many unanswered questions remain. In particular, the underlying cause for the anomaly in the case-fatality ratio is unclear. The clear example is Mexico, which is the 13[th] country with the highest confirmed cases, however, with the 4[th] highest mortality. Is this anomaly due to interrelationship between disease and factors related to the environment/geographical location? The notion that COVID-19 mortality varies locally forms the basis of this study.

This study aims to explore the spatial distribution of COVID-19 mortality and relationship to country and local level data on demographics and pre-existing health conditions. To address the central question, two databases are used: a) a publicly available country aggregate dataset of global COVID-19 cases to explore the position of Mexico amongst the other countries [3] and a dataset provided by the Mexican Ministry of Health. The latter is a unique patient-level dataset detailing health background, demographics and associated comorbidities for each COVID-19 patient together with disease outcome [4].

## 2 State of the Art

One of the earliest studies (February 2020) to investigate the spatial analysis of the current pandemic used data from ~ 1100 COVID-19 patients together with their demographic data including gender and age to identify the spread of the virus through Chinese provinces, whether they experienced contact with wild-life or visited Wuhan, the currently reported origin of the pandemic. The authors mapped the distribution of patients across Mainland China [5]. Furthermore, *Chen et al.*, (2020), examined the spatial distribution of COVID-19 patients and correlation with the migration of the Wuhan resident during the early phase of the crisis and concluded that Wuhan residents were the main source of infection spread to other Chinese provinces [6]. In addition, using spatiotemporal analysis, Huang *et al.*, (2020) reported that a high number of COVID-19 cases in Wenzhou province (neighbouring pandemic origin) in earlier phase was due to a significant number of residents travelling from Wuhan to Wenzhou [7].

Other studies have examined the relationship between COVID-19, geography and/or health and socioeconomic factors. For example, Padula and Davidson (2020) investigated the country-based relationship between COVID-19 mortality and number of healthcare workers and reported an inverse relationship [8]. Moreover, Mollalo *et al.*, (2020), studied the relationship between a range of socioeconomic, behavioural, environmental and demographic factors of COVID-19 patients to their geographical locations [9]. The authors reported that income, number of healthcare workers in the area and ethnicity correlate with a high mortality rate in the US states. Other studies mapped COVID-19 data to geographical locations in Brazil [10], India [11], USA [12], Spain [13], Iran [14], Israel [15], Mexico [16] and Pakistan [17].

These studies have significantly enhanced our understanding of the epidemiological spread of the disease and provided spatial information, typically by a map with the weight of the reported measure (mortality, survival, etc.) represented by colour intensity or symbols of different sizes [12].

This approach, nonetheless, leads to a visual bias by over-emphasizing geographically large but sparsely populated area, conversely underrepresenting densely populated cities. This is particularly a significant issue as the majority of COVID-19 mortality has occurred within densely populated large cities.

An alternative approach is by assigning the same weight to each area for equal representation. One method to achieve this is by hexbin cartogram, a new method that is gaining popularity particularly for election reporting [18]. This allows revealing patterns in conventional choropleth maps that are otherwise hidden due to geographical weight of such maps. Using hexbin plots Beecham *et al.*, 2018 reported metropolitan weight of electorate that voted Remain in the 2016 UK referendum on membership of the European Union.

Herein, the spatial distribution and relationship of COVID-19 mortality in Mexico, as one of the most affected countries, will be examined. In addition, the spatial relationship of mortality to a range of pre-existing health conditions and demographics (age, sex) will be investigated at country and local levels.

## 3   Properties of the Data

Two datasets are used in this study: a) country aggregate data provided by the Lancet COVID-19 Commission [3] to visualise the position of Mexico amongst the affected countries and; b) Mexico COVID-19 patient-level data released by the Mexican Ministry of Health [4] for spatial mapping.

The country aggregate data contain details of COVID-19 in 201 countries from the start of the pandemic. It currently contains 4.8 million datapoints covering 201 countries, 374 days with 154 indicators related to cases confirmed, mortality, demographics pre-existing health and socioeconomic status as well as measures related to stringency levels of government response around the world. Measures were taken to clean this dataset reported recently [19]; briefly, the missing values were removed and a subset of data related to the top 15 affected countries used to plot confirmed cases, recovered and mortality. In addition, time-series data for recovery and mortality in Mexico from April/2020 to Nov/2020 were extracted to plot and identify any potential pattern.

Mexico COVID-19 data [4] detailing individual-level information of patients and their pre-existing health background (e.g. obesity, diabetes, pulmonary heart disease, chronic kidney disease, etc.), behavioural (smoking), demographics (age, gender) and outcome of the disease and whether the patient survived or deceased. The dataset contains information related to 3,628,951 patients with 40 columns. Using the dictionary provided the name of the columns renamed from Spanish to English and the variables unrelated to the remit of this study (e.g., record ID) were removed. Missing and partial data related to patients were also removed. The distribution of variables used for analysis is plotted and visualised in Figure1. Data show that the majority of COVID-19 patients survived, further analysis revealed that 175,494 out of total 3,628,951 patients deceased. Survival data removed and analysis focused on mortality.

Information in this dataset is in a binary format (1: True, 2: False), therefore, to obtain the count of patients for each Mexican state and municipalities within, a series of grouping operations were performed. This resulted in a table containing data related to mortality only with the total count of males, females, whether patients intubated, exhibited pneumonia, diabetes, had the chronic obstructive pulmonary disease (COPD), asthmatic, immunosuppression, hypertension, cardiovascular disease, obesity, chronic kidney disease and age ranges of 0-20, 20-40, 40-60, 60-80 and 80-110. These operations were performed in Python.

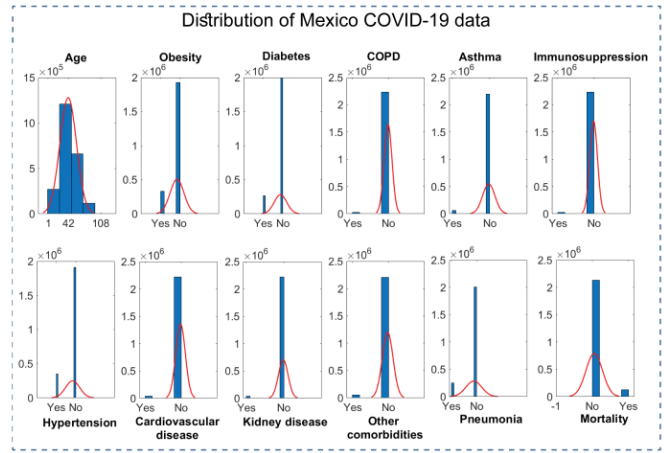

Fig. 1. Distribution of Mexico COVID-19 data. Variables plotted include age, obesity, diabetes, chronic obstructive pulmonary disease (COD), asthma, immunosuppression, hypertension, cardiovascular and kidney disease, presence of other comorbidities, pneumonia and mortality.

The resulting dataset imported into R and merged with a sample dataset with "mxmaps" library to obtain a population of each state and municipality as well as to generate the dataset required by the library for plotting thematic choropleth and hexbin maps [20].

## 4   Analysis

### 4.1   Approach

The steps which are taken in this study detailed in Figure 2. Initially, the status of COVID-19 mortality in Mexico amongst other countries explored. To achieve this, bar plots of confirmed cases, recovered, and mortality plotted using Seaborn library for Python. Subsequently, the number of recovered and mortality per million/population plotted to visually examine patterns.

The central tenet in this study is to examine spatial variation in COVID-19 mortality in Mexico states and local municipalities. If found, these associations will be further examined at a local level.

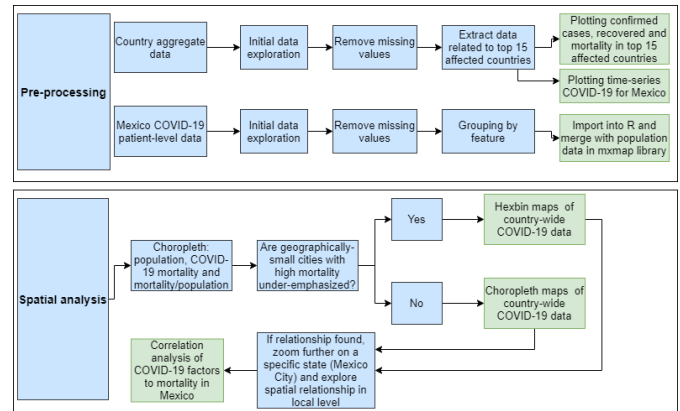

Fig. 2. Workflow for the analysis of COVID-19 data performed in this study.

To ascertain geographical link to disease it is important to present data without inherent bias. This is particularly an issue when a country of interest is composed of a large sparsely populated and small densely populated area. A clear example in Mexico is the comparison between the large state of Chihuahua (247,455 square km) on the north of Mexico with a population of 3,376,062 and mortality of 6,431. In contrast, Mexico City, a geographically small (1,485 square km) is a densely populated city (8,918,653) and the 32nd Mexican state that has suffered 22,880 mortality. Whilst, the geographical size of Chihuahua is ~167 times bigger than Mexico City, its population is 260% smaller than Mexico City with a reported 355% less mortality.

The approach in this study to overcome this bias is to use a method of equal representation of each geographical area. This is indeed used in spatial analysis of elections where analysts are faced with a similar dilemma. Hexbin mapping produces a map similar to a scatter plot, with the graphing space divided into equally-sized hexagonal cartograms. A variable such as the number of voters can subsequently be represented on this hexagons by colour intensity [18]. This is an improvement; however, it is a newly reported methodology and as such readers might not be able to immediately recognise corresponding geographical area to their respective hexagons.

The hexbin mapping in this study is achieved by using "mxmaps" R library [20] which is based on "choroplethr" package. It is also able to use "leaflet" package to create interactive maps. The mxmaps library contains sample dataset which includes population per state and municipality. The Mexico data is merged in R with the sample dataset to integrate population data. For comparison, the choropleth maps together with the hex maps of the entire country are produced using RColorBrewer reversed RdYlBu colouring palette. This is followed by spatial state mapping of data and subsequently spatial choropleth mapping of data in Mexico City.

The correlation of pre-existing health conditions and demographics to mortality is further investigated by correlation matrix. Finally, dimensionality reduction is employed to select predictors that can optimally model the disease outcome. This is achieved by lassoglm in MATLAB function on train and test datasets. Spatial analysis will inform whether dividing the dataset into male and females is more appropriate for this step.

### 4.2 Process

#### 4.2.1 Analysis of country aggregate data

Initially, the position of Mexico amongst the top 15 affected countries is explored. Our data show that Mexico is the 13th country for the number of COVID-19 confirmed cases with 1,437,185 and the US being the 1st with 20,132,054 cases. Mexico occupies 10th place on the recovery chart with 1,083,768 and India having the highest recovery at 9,883,461. Mexico has the 4th highest mortality with 126,507 and the US is at the top with 347,894 mortality (Fig. 3A). The position of Mexico on the mortality chart is not on par with its position on the number of cases and recovery.

Is this anomaly on the ratio of COVD-19 cases and mortality due to the unusual pattern in mortality during the pandemic? Analysis of recovery and mortality over an 8 month period indicate that whilst a peak in recovery between July and August of 2020 is observed, that the number of mortality normalised per million of the population has been consistently below 10 with one exception of 23/million in October 2020 (Fig.3B).

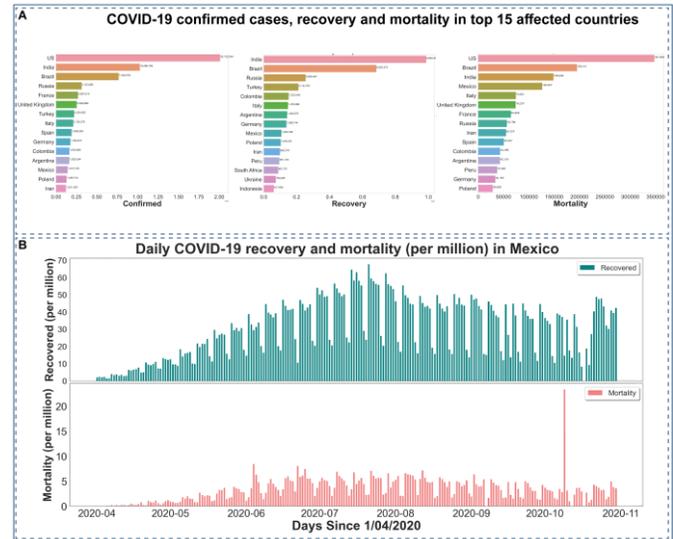

Fig. 3. COVID-19 data in top 15 affected countries and Mexico. A) Confirmed cases, recovery and mortality in top 15 affected countries and B) daily COVID-19 recovery and mortality per million in Mexico.

Is this consistently high case/mortality ratio better explained by spatial analysis of locally-influenced factors?

#### 4.2.2 Traditional choropleth for COVID-19 mapping?

Is traditional choropleth the best method to visually convey COVID-19 spatial mapping in Mexico? This study sought to test this question by comparing choropleth and hexbin maps. Our data show that whilst traditional choropleth maps retain geographical proportionality, they, nevertheless, are unable to visually reveal the extent of the crisis.

Figure 4A shows that the size of 32 Mexican States has significant variations. For example, the northern states are less populated and occupy a significantly larger area than central states which are small densely populated. Spatial mapping of absolute mortality and mortality normalised/population of each state reveal a similar trend. Two central states named Mexico (this is also the name of a state, in addition to the country named Mexico and Mexico City as Capital) and Mexico City together account for ~32% of total country-wide mortality, however not visible according to their weight. This comparison is highlighted in Figure 4A-B.

Taken together, our data show that hexbin maps are a more appropriate method to spatially analyse COVID-19 data country-wide and thus this will be used in further analysis of country-wide data.

#### 4.2.3 Spatial mapping of sex and age-related differences in COVID-19 mortality in entire Mexico

Is there a difference between males and females COVID-19 mortality in Mexico and is this difference revealed spatially? Our data show that males account for 63% of total mortality whilst females for 37%. This can not be explained by the slight difference in the proportion of males and females (48.6% and 51.4% of the total population for males and females respectively). Hexbin mapping of mortality separated by sex, show that male mortality is higher in 7 States and lower in 6 than females (Fig. 5A).

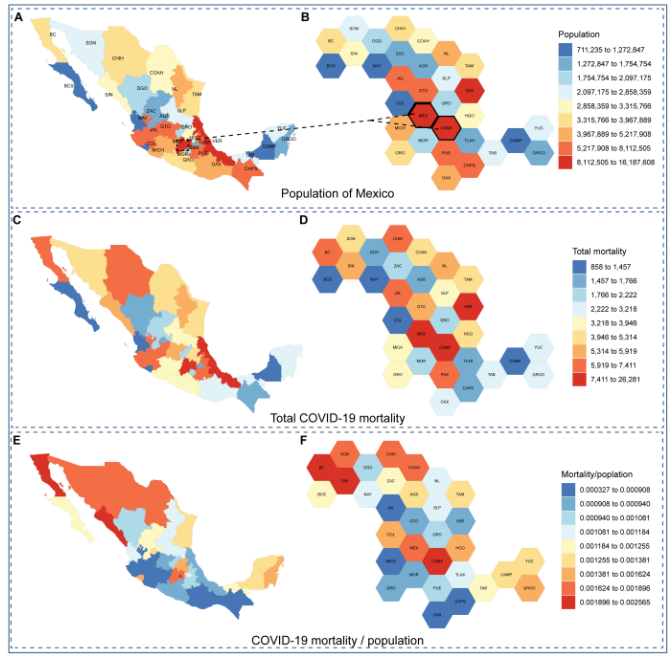

Fig. 4. Spatial mapping of Mexico population, COVID-19 mortality and mortality/population using traditional choropleth (A, C and E) and hexbin maps (B, D and F). The dashed box in A show two comparable States accounting for ~32% of COVID-19 mortality using choropleth and hexbin mapping.

Further analysis of mortality by age range mapped both as a proportion of total mortality and normalised against the population, revealed spatial divergence with age range. As a proportion to total mortality, more patients at age range 0-20 deceased in the southern and northern states with the central states including Mexico City suffering the least proportion of 0-20 mortality. With regard to the age range 20-40 and 40-60 more patients deceased in central states including Mexico City.

A trend was also observed for 60-80 and 80-110 years of age with more patients deceased in the northern states. Plotting of the same data, normalised against the total population show a slightly different trend. The northern states and the central state of Mexico and Mexico City (except for 0-20) suffered high mortality consistently across all age ranges. The southern states account for higher mortality for 0-20 but the lowest for the other age ranges (Fig. 5B).

### 4.2.4 Spatial mapping of pre-existing conditions in deceased COVID-19 patients in entire Mexico

To what extent these spatial variations are explained by pre-existing health conditions. To answer this question, patient-related data (diabetes, obesity, cardiovascular disease, hypertension, asthma, chronic kidney disease) and two COVID-19 related clinical features (pneumonia and intubation) were plotted using hexbin maps. Our data show a spatial trend for deceased patients with pre-existing conditions. For example, except diabetes, Mexico City has consistently the highest deceased patients with obesity, cardiovascular disease, hypertension, asthma and kidney disease. Similarly, more deceased patients in Mexico City suffered from pneumonia and were intubated. Southern states (e.g. Oaxaca and Campeche) exhibited (except for diabetes) the least number of deceased patients with pre-existing conditions (Fig. 6).

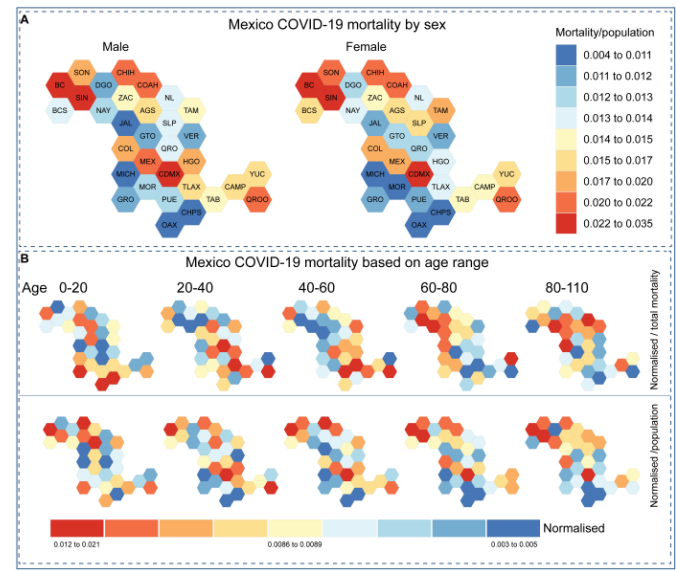

Fig. 5. Hexbin mapping of COVID-19 mortality in Mexico separated by sex (A) and age (B).

### 4.2.5 Spatial mapping of age-related differences in COVID-19 mortality in Mexico City

Are these nationally-divergent observations also evident at the local level? To address this question, a subset of data related to Mexico City extracted and plotted using traditional choropleth mapping. Our data show that whilst, the eastern municipality of Iztapalapa is the most populated and has the highest absolute number of mortality, once this is normalised to the local population it no longer retains the highest mortality. The second most populated municipality in Mexico City (Fig.7A top: Gustavo Madero) has suffered the second-highest absolute number and remains the second-highest after normalisation. The northern municipality of Azcapotzalco is sparsely populated but has the highest normalised mortality. In contrast, the most southern municipality (Fig.7A bottom: Milpa Alta) is sparsely populated and has experienced the lowest number of COVID-19 mortality.

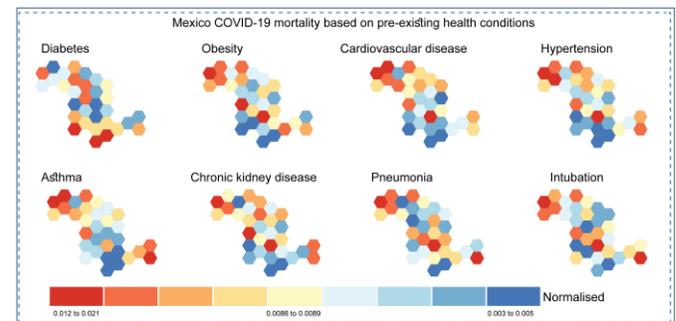

Fig. 6. Mexico COVID-19 spatial mapping of pre-existing conditions amongst deceased patients. These conditions are: diabetes, obesity, cardiovascular disease, hypertension, asthma and chronic kidney disease. In addition, two COVID-19 related clinical outcomes (pneumonia and intubation) are also plotted.

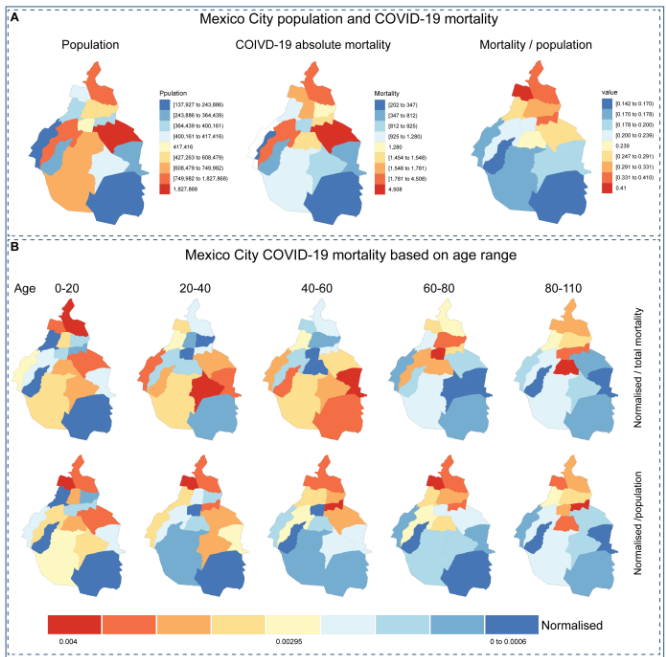

Fig. 7. Mexico City mapping of population, absolute COVID-19 mortality and mortality normalised by population (A) and age range of deceased COVID-19 patients mapped onto municipalities within the City.

Further break down of mortality by age range normalised both against the total mortality and local population reveal that the northern municipalities of Azcapotzalco and Gustavo Madero have suffered the highest proportion of mortality. In contrast, the southern municipality of Milpa Alta has experienced the lowest mortality for 0-20, 20-40, 60-80 and 80-110. Normalised against the total mortality this municipality suffered high mortality amongst 40-60 years of age, this, however, was not apparent when normalised against the population (Fig. 7).

### 4.2.6 Spatial mapping of pre-existing conditions in deceased COVID-19 patients in Mexico City

Is the country-wide trend for the association of pre-existing conditions to mortality apparent at local Mexico City level?

Our analysis shows a clear spatial pattern of pre-existing conditions amongst deceased patients in Mexico City. Four northern municipalities of Azcapotzalco, Gustavo Madero, Venustiano Carranza and Iztacalco have the highest diabetes, obesity, cardiovascular disease, hypertension, asthma and kidney disease. The deceased in these municipalities were most likely to suffer from pneumonia and to be intubated (Fig. 8).

Similar to its position on the breakdown of age-related mortality in Mexico City, the municipality of Milpa Alta again showed the lowest values for pre-existing conditions (Fig. 8).

### 4.2.7 Is COVID-19 mortality in Mexico dependent on pre-existing conditions?

Spatial analysis revealed an association of mortality both at the country and local level. These visual findings have informed and led to further statistical analysis.

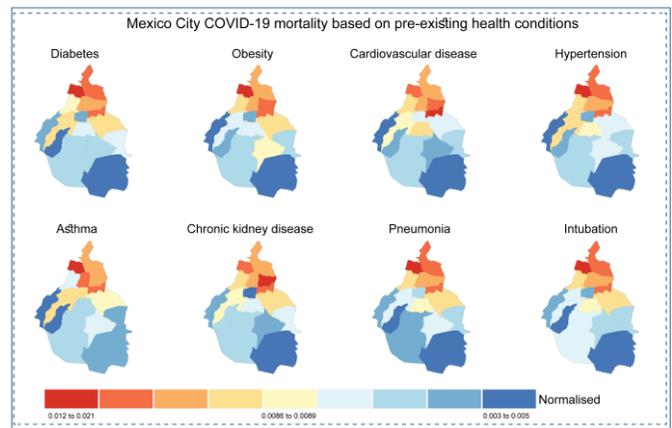

Fig. 8. Mexico City COVID-19 spatial mapping of pre-existing conditions amongst deceased patients. These conditions are: diabetes, obesity, cardiovascular disease, hypertension, asthma and chronic kidney disease. In addition, two COVID-19 related clinical outcomes (pneumonia and intubation) are also plotted.

Correlation analysis of pre-existing health conditions and demographics including age and gender revealed that in both male and females, pneumonia, hypertension and diabetes were negatively related to survival. Age was positively correlated with mortality (Fig. 9). To ascertain whether these variables could contribute to the prediction of mortality, lasso regularisation was performed. This analysis showed that for males, age; obesity; diabetes; immunosuppression; hypertension; kidney disease; other comorbidities and pneumonia are significant contributors to disease outcome. In addition, for females, age; obesity; diabetes; chronic obstructive pulmonary disease, immunosuppression, hypertension, kidney disease and other comorbidities are important predictors for disease outcome.

### 4.3    Results

The main findings of this study are that:

a)  Mexico suffers from COVID-19 case/mortality ratio imbalance. Whilst it occupies the 13[th] place on the list of countries with the highest confirmed cases, its place on the mortality list is 4[th]. The rate of mortality per million has been consistent for the past several months.

b)  The use of hexagonal cartograms is a better approach for spatial mapping of COVID-19 data in Mexico as it addresses bias in area size and population.

c)  Sex and age-related spatial relationship with mortality is observed amongst the Mexican states.

d)  There is a spatial trend between pre-existing health conditions and mortality at the state level.

e)  Within Mexico City, there is a clear south, north divide with higher mortality in the northern municipalities.

f)  Deceased patients in these northern municipalities have the highest pre-existing health conditions.

g)  Correlation analysis revealed that all factors were correlated with the disease outcome each at different levels with age, diabetes, hypertension and pneumonia having the strongest correlation.

h) Lasso regularization and feature selection showed that aspects of pre-existing health conditions are strong predictors of disease outcome.

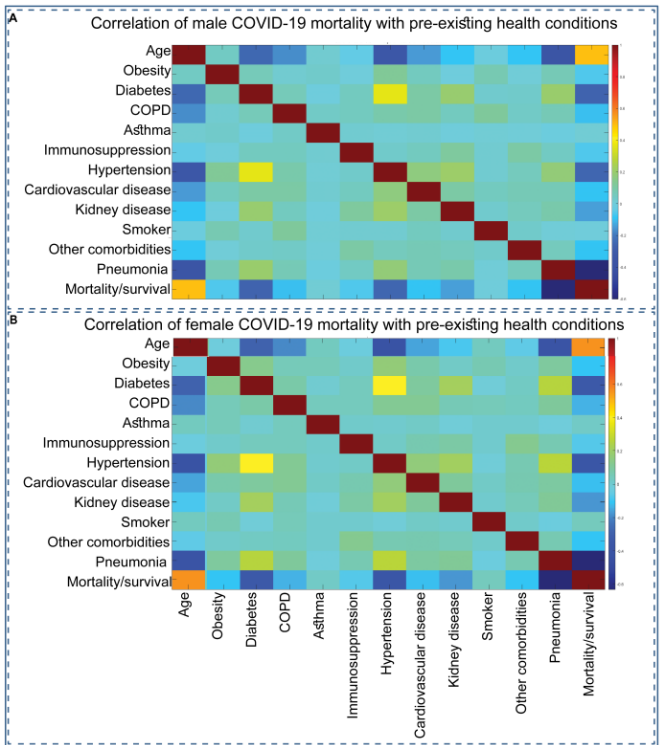

Fig. 9. COVID-19 data in top 15 affected countries and Mexico. A) Confirmed cases, recovery and mortality in top 15 affected countries and B) daily COVID-19 recovery and mortality per million in Mexico.

## 5 CRITICAL REFLECTION

In this study, a predominantly descriptive, spatially centred examination of COVID-19 mortality and related factors at both country and local levels is presented.

The central aim was to achieve this spatial mapping without visual bias, inherent in presenting data related to densely packed small regions vs. sparsely populated large area. Within this central aim, several sub-aims were explored: i) to investigate the spatial relationship between COVID-19 mortality, demographics and patient background at country level; ii) to examine these relationships at the local municipality level, and iii) to test for correlation and to identify important features for disease outcome.

In agreement with the newly described approach employed to report election data [18], findings from this study indicate that hexagonal cartograms are better suited to report COVID-19 data in Mexico. This study reports spatial patterns of mortality in the Mexican states with the most southern states least affected by the pandemic except for young patients between 0-20. The central/northern states are affected to a greater degree with similar pattern for pre-existing conditions.

Further exploration of these observations for Mexico City, revealed clear patterns for the spatial distribution of mortality Mexico City is a megacity that with the surrounding area accounts for about 20 million inhabitants [21]. Due to its rapid expansion, appropriate infrastructure including healthcare provisions not built accordingly [22, 23]. This large concentration of residents with significant pre-existing health complications has led for Mexico City and neighbouring Mexico state to account for about 1/3 of total COVID-19 mortality. Our data, in agreement with a previous report, show that Mexico City has the highest pre-existing health conditions including diabetes, hypertension and obesity. This high prevalence is apparent amongst young and old similarly [24].

This study has several limitations. These include exclusion of patients that survived, a discrepancy between reported fatality by the Mexican Ministry of Health (154,441) and John Hopkins University (133,204). It is likely that data provided by the former to be accurate, however, these need to be explored. In addition, the basis for these spatial relationships needs to be further explored as to why within Mexico City the mortality are concentrated in particular regions. Inclusion of locally-derived factors related to socioeconomic and health background may, to an extent, address this shortcoming. Furthermore, correlation analysis, feature selections and prediction of outcomes should consider these local variations and build in measures to account for this variability.

Adequate inferences from spatial analysis of COVID-19 data may allow better understanding for the management of highly infectious disease amongst densely populated regions. This approach is not new as many previous studies during the current and previous pandemics were performed [25-29]. There is, however, scope for improvement in visualisation.

Taken together, this study provides an improved presentation of COVID-19 mapping in Mexico and demonstrates spatial divergence of the mortality in Mexico.

## REFERENCES

[1] C. Huang *et al.*, "Clinical features of patients infected with 2019 novel coronavirus in Wuhan, China," *The lancet,* vol. 395, no. 10223, pp. 497-506, 2020.

[2] J. Hopkins, "Johns Hopkins Coronavirus Resource Center," *COVID-19 Case Tracker,* 2020.

[3] J. D. Sachs, R. Horton, J. Bagenal, Y. B. Amor, O. K. Caman, and G. Lafortune, "The Lancet COVID-19 Commission," *The Lancet,* vol. 396, no. 10249, pp. 454-455, 2020.

[4] M. M. o. Health. *Mexico COVID-19 dataset*. [Online]. Available: https://github.com/alberto-mateos-mo/covid19mx

[5] W.-j. Guan *et al.*, "Clinical characteristics of coronavirus disease 2019 in China," *New England journal of medicine,* vol. 382, no. 18, pp. 1708-1720, 2020.

[6] Z.-L. Chen *et al.*, "Distribution of the COVID-19 epidemic and correlation with population emigration from Wuhan, China," *Chinese medical journal,* 2020.

[7] H. Huang *et al.*, "Epidemic Features and Control of 2019 Novel Coronavirus Pneumonia in Wenzhou, China," *China (3/3/2020),* 2020.

[8] W. V. Padula and P. Davidson, "Countries with High Registered Nurse (RN) Concentrations Observe Reduced Mortality Rates of Coronavirus Disease 2019 (COVID-19)," *Available at SSRN 3566190,* 2020.

[9] A. Mollalo, B. Vahedi, and K. M. Rivera, "GIS-based spatial modeling of COVID-19 incidence rate in the continental United States," *Science of The Total Environment,* p. 138884, 2020.

[10] R. Dagnino, E. J. Weber, and L. M. Panitz, "Monitoramento do Coronavírus (Covid-19) nos municípios do Rio Grande do Sul, Brasil," *SocArXiv. March,* vol. 28, 2020.

[11] A. Saha, K. Gupta, and M. Patil, "Monitoring and Epidemiological Trends of Coronavirus Disease (COVID-19) Around The World," 2020.

[12] E. Dong, H. Du, and L. Gardner, "An interactive web-based dashboard to track COVID-19 in real time," *The Lancet infectious diseases,* vol. 20, no. 5, pp. 533-534, 2020.

[13] L. Orea and I. C. Álvarez, "How effective has the Spanish lockdown been to battle COVID-19? A spatial analysis of the coronavirus propagation across provinces," *Documento de Trabajo,* p. 03, 2020.

[14] Z. Arab-Mazar, R. Sah, A. A. Rabaan, K. Dhama, and A. J. Rodriguez-Morales, "Mapping the incidence of the COVID-19 hotspot in Iran–Implications for Travellers," *Travel Medicine and Infectious Disease,* 2020.

[15] H. Rossman *et al.*, "A framework for identifying regional outbreak and spread of COVID-19 from one-minute population-wide surveys," *Nature Medicine,* vol. 26, no. 5, pp. 634-638, 2020.

[16] M. S. Juárez, "COVID-19 en México: comportamiento espacio temporal y condicionantes socioespaciales, febrero y marzo de 2020," *Posición,* vol. 3, pp. 2683-8915, 2020.

[17] S. Sarwar, R. Waheed, S. Sarwar, and A. Khan, "COVID-19 challenges to Pakistan: Is GIS analysis useful to draw solutions?," *Science of The Total Environment,* p. 139089, 2020.

[18] R. Beecham, A. Slingsby, and C. Brunsdon, "Locally-varying explanations behind the United Kingdom's vote to leave the European Union," *Journal of Spatial Information Science,* vol. 2018, no. 16, pp. 117-136, 2018.

[19] B. Javaheri, "The COVID pandemic: socioeconomic and health disparities," *arXiv preprint arXiv:2012.11399,* 2020.

[20] D. Valle-Jones. "mxmaps - create maps of Mexico." https://github.com/diegovalle/mxmaps/ (accessed.

[21] C. Paquette, "OECD Territorial Reviews: Valle de México, Mexico," ed: Editions de l'OCDE, 2015.

[22] B. Tellman *et al.*, "Adaptive pathways and coupled infrastructure: seven centuries of adaptation to water risk and the production of vulnerability in Mexico City," *Ecology and Society,* vol. 23, no. 1, 2018.

[23] A. Baeza, A. Estrada-Barón, F. Serrano-Candela, L. A. Bojórquez, H. Eakin, and A. E. Escalante, "Biophysical, infrastructural and social heterogeneities explain spatial distribution of waterborne gastrointestinal disease burden in Mexico City," *Environmental Research Letters,* vol. 13, no. 6, p. 064016, 2018.

[24] L. Parra-Rodríguez *et al.*, "The burden of disease in Mexican older adults: premature mortality challenging a limited-resource health system," *Journal of aging and health,* vol. 32, no. 7-8, pp. 543-553, 2020.

[25] B. Meng, J. Wang, J. Liu, J. Wu, and E. Zhong, "Understanding the spatial diffusion process of severe acute respiratory syndrome in Beijing," *Public Health,* vol. 119, no. 12, pp. 1080-1087, 2005.

[26] L. Q. Fang *et al.*, "Geographical spread of SARS in mainland China," *Tropical Medicine & International Health,* vol. 14, pp. 14-20, 2009.

[27] K. Al-Ahmadi, S. Alahmadi, and A. Al-Zahrani, "Spatiotemporal clustering of Middle East respiratory syndrome coronavirus (MERS-CoV) incidence in Saudi Arabia, 2012–2019," *International journal of environmental research and public health,* vol. 16, no. 14, p. 2520, 2019.

[28] Q. Lin, A. P. Chiu, S. Zhao, and D. He, "Modeling the spread of Middle East respiratory syndrome coronavirus in Saudi Arabia," *Statistical methods in medical research,* vol. 27, no. 7, pp. 1968-1978, 2018.

[29] O. A. Adegboye, E. Gayawan, and F. Hanna, "Spatial modelling of contribution of individual level risk factors for mortality from Middle East respiratory syndrome coronavirus in the Arabian Peninsula," *PloS one,* vol. 12, no. 7, p. e0181215, 2017.